\documentclass[sigconf]{acmart}
\setcopyright{rightsretained}
\copyrightyear{2021}
\acmYear{2021}
\acmDOI{}

\acmConference[KDD MLF '21]{KDD MLF '21: KDD Workshop on Machine Learning in Finance}{August 15, 2021}{Virtual}
\acmPrice{}
\acmISBN{}

\usepackage{cleveref}
\usepackage{mathtools} 
\newtheoremstyle{sltheorem}
{}                % Space above
{3 pt}             % Space below
{\slshape}        % Theorem body font % (default is "\upshape")
{}                % Indent amount
{\bfseries}       % Theorem head font % (default is \mdseries)
{.}               % Punctuation after theorem head % default: no punctuation
{ }               % Space after theorem head
{}                % Theorem head spec
\theoremstyle{sltheorem}

\newtheorem{challenge}{Challenge}
\begin{document}
\title{Seven challenges for harmonizing explainability requirements}

\author{Jiahao Chen}
\email{jiahao.chen@jpmchase.com}
\orcid{0000-0002-4357-6574}
\affiliation{%
  \institution{J.\ P.\ Morgan AI Research}
  \streetaddress{383 Madison Avenue}
  \city{New York}
  \state{New York}
  \country{USA}
  \postcode{10179-0001}
}
\author{Victor Storchan}
\email{victor.storchan@jpmorgan.com}
\affiliation{%
  \institution{J.\ P.\ Morgan}
  \streetaddress{310 University Avenue}
  \city{Palo Alto}
  \state{California}
  \country{USA}
  \postcode{94301-1744}
}

% \renewcommand{\shortauthors}{Chen et al.}

\begin{abstract}
Regulators have signalled an interest in adopting explainable AI (XAI)
techniques to handle the diverse needs for model governance,
operational servicing, and compliance in the financial services industry.
In this short overview, we review the recent technical literature in XAI
and argue that based on our current understanding of the field,
the use of XAI techniques in practice necessitate a highly contextualized
approach considering the specific needs of stakeholders
for particular business applications.
\end{abstract}


% \keywords{datasets, neural networks, gaze detection, text tagging}


\maketitle

\section{Motivation}
Artificial intelligence (AI) systems are ubiquitous components
of financial, compliance and operational business processes such as
marketing \citep{Jannach2019},
credit decisioning \citep{Hurley2016},
client satisfaction \citep{Zhao2019},
cybersecurity \citep{Siddiqui2019}, and
anti-money laundering \citep{Weber2019}.
The growing complexity of these systems pose
new challenges for
for model governance \citep{Kurshan2020} 
fair lending compliance \citep{Chen2018},
and other regulatory needs.

However, federal regulation in the 
United States of America (US)
has yet to catch up to the frenetic pace of innovation in AI
systems, both in research and practical deployments.
For example, the OCC 2011-12 / Federal Reserve SR 11-7
Guidance on Model Risk Management \citep{modelrisk},
published in 2011,
is now ten years old and predates
the current explosion of deep learning
in AI.
Similarly, federal fair lending laws like the Equal Credit Opportunity Act (ECOA,
a.k.a.\ Regulation B; enacted in 1974)
mandate a customer's right to explanation when financial institutions take actions that
adversely impact their access to credit,
and other laws like the Fair Credit Reporting Act (FCRA; enacted in 1970)
protect a customer's right to dispute incorrect data on their credit reports \citep{Chen2018}.

The age of financial regulations have hindered regulatory enforcement.
A recent investigation from the New York Department of Financial Services
into alleged fair lending discrimination in the Apple Card credit card 
concluded that the letter of fair lending law was not violated
insomuch as alleged spousal inequities were fully attributable to different data in the
spouses' credit reports \citep{applecard},
while at the same time noting that
``the use of credit scoring in its current form and laws and regulations barring discrimination in lending are in need of strengthening and modernization to improve access to credit'' \citep{nydfspr}. 
Recently, federal regulators have issues a joint Request for Information
on the use of AI in financial services \citep{usrfi},
signalling an interest to refresh financial regulations
in the light of the Biden Administration's focus on AI \citep{Samp2021}.

\paragraph{Contributions}
In \Cref{sec:challenges},
we review the recent literature on explainable AI (XAI)
and organize the findings into
seven major challenges that pose practical considerations
when specifying needs for XAI in industry.
We conclude in \Cref{sec:conclusions} that the
diversity of core AI technologies, stakeholders and
potential applications in the industry,
when coupled with the relative immaturity of XAI tools,
necessitate an intent-based approach to 
match stakeholder needs to the most appropriate
XAI technique in the context of specific use cases.

\section{The challenges}\label{sec:challenges}

\begin{challenge}
The intent and scope of explanations matter.
\end{challenge}

Different stakeholders have different needs for explanation \citep{Tomsett2018,Bhatt2020},
but these needs are not often well-articulated or distinguished from each other
\citep{Miller2019,Ras2018,Xie2020,Kaminski2020,Kasirzadeh2019}.
Clarity on the intended use of explanation is crucial to select an appropriate XAI tool,
as specialized methods exist for specific needs like
debugging \citep{Kang2018},
formal verification (safety) \citep{Cousot1992,Guerra2009,Xu2020},
uncertainty quantification \citep{Wang2020,Abdar2020},
actionable recourse \citep{Ustun2019,Karimi2020},
mechanism inference \citep{Deva2021},
causal inference \citep{Pearl2009,Frye2019,Bertossi2020},
robustness to adversarial inputs \citep{Madry2018,Lecuyer2019},
data accountability \citep{Yona2021},
social transparency \citep{Ehsan2021b},
interactive personalization \citep{Virgolin2021},
and
fairness and algorithmic bias \citep{Ntoutsi2020}
.
In contrast, feature importance methods like LIME \citep{lime}
and SHAP \citep{Lundberg2017,Lundberg2018}
focus exclusively on computing quantitative evidence for indicative conditionals \citep{Harper1981,Bennett2003}
(of the form ``If the applicant doesn't have enough income, then she won't get the loan approved''),
with some newer counterfactual explanation methods \citep{Artelt2019,Mothilal2020,Stepin2021}
and negative contrastive methods \citep{Luss2019}
finding similar evidence for subjunctive conditionals \citep{Pollock1976,Byrne2016}
(of the form ``If the applicant increases her income, then she would get the loan approved'')
.

\begin{challenge}
The type of data and type of model matter.
\end{challenge}

In addition, the type of explanation changes depending on the data source
such as images \citep{Kindermans2019,Zhu2021},
symbolic logic representation \citep{Albini2020},
or
text based explanations \citep{Amir2018}
.
Yet other methods exist for domains beyond supervised learning / classification,
applied instead to
unsupervised learning / clustering \citep{Horel2019,Sarhan2019,Dasgupta2020},
reinforcement learning \citep{Madumal2020},
AI planning \citep{Cashmore2019},
computer vision \citep{Zhu2021},
recommendation systems \citep{Zhang2020}, 
natural language processing \citep{Mullenbach2018,Yang2019,Wawrzinek2020},
speech recognition \citep{Hsu2018},
or
multi-agent simulations \citep{Amir2018}
.
Specialized XAI techniques even exist for adaptive systems such as
interactive visualization systems \citep{Vig2021},
interactive virtual agents \citep{Kambhampati2019,Weitz2019},
active learning \citep{Ghai2020},
and human-in-the-loop systems \citep{Yang2019}
.
Furthermore, comparative studies across multiple XAI techniques have shown low mutual agreement \citep{Neely2021},
which is consistent with internal research findings at AI Research.
With these research findings taken together,
we expect that different explanation tools will be needed to address each problem domain effectively.

\begin{challenge}
The human factors around intent and scope of explanations matter.
\end{challenge}
Results from psychology and other social sciences highlight the social nature of explanation
as a critical design criterion \citep{Miller2019}.
Some authors like \citet{Kumar2020} have argued that Shapley value-based explanations do not satisfy
human needs for explanation, in particular, the desire to read causality from the explanations.
Understanding human-centric factors in explanations for diverse stakeholders
is now an area of intense research \citep{Ehsan2020,Ehsan2021} and we expect new relevant results
to emerge rapidly.
In particular, a recent paper from Microsoft Research \citep{Kaur2020} provides worrying
results that humans generally over-trust AI-generated explanations
regardless of their level of subject matter expertise.
A detailed human--computer interaction (HCI) study of the use of XAI
techniques for explaining fraud models did not show
uniform superiority of any one explanation technique \citep{Jesus2021}.

\begin{challenge}
There is no consensus around evaluating the correctness of explanations.
\end{challenge}
One of the biggest impediments to practical consumption of XAI
is the inherent difficulty to evaluate if a given explanation is correct or not.
For \textit{post hoc} explanation methods, the usual metric is \textit{fidelity},
namely how accurately the surrogate model built by the explanation method
approximates the true model \citep{Plumb2018}.
In fact, LIME is explicitly constructed around a trade-off between fidelity
and the complexity of the resulting explantion \citep{lime};
other recent work generalizes the purpose of explanations as a multiobjective game
with trade-offs between explanatory accuracy, simplicity and relevance \citep{Watson2020}.
However, fidelity is measured over the entire possible domain of inputs to a model
and therefore places undue emphasis on unobserved and infeasible parts of the input space
\citep{Frye2019}.
Apart from such quantitative assessments of correctness,
which are not free of problems,
the best we can do is appeal to formal philosophical notions of
epistemology \citep{Mittelstadt2019},
but at the cost of any quantification of explanatory accuracy.

\begin{challenge}
XAI techniques in general lack robustness and have strong basis dependence.
\end{challenge}
Local feature importance methods SHAP and LIME are known to not be robust \citep{Alvarez-Melis2018,Slack2020}.
Several studies have demonstrated that Shapley-value based methods like SHAP suffer from an effect similar in spirit to multicollinearity,
assigning spuriously low importances to highly dependent features \citep{Hooker2019}.
In the extreme case of identical features, the result Shapley values is effectively averaged
over each feature, thus resulting in artificially lowered feature importances \citep{Kumar2020}.
In addition, there are subtle dependencies of Shapley value-based methods on the data distribution \citep{Frye2019,Chen2020,Sundararajan2020}.
Similar results exists for other techniques such as influence functions, which arise from nonconvex effects \citep{Basu2021}.
Such results caution that the basis of features has to be carefully considered in crafting
an explanation.

\begin{challenge}
Feature importance explanation methods can be manipulated.
\end{challenge}
Recent work on ``fairwashing'' has demonstrated that feature importance methods can be
easily manipulated to create arbitrary feature importance rankings,
by artificially varying the behavior of the model on unobserved regions of the input space \citep{Dombrowski2019,Aivodji2019,Anders2020,Slack2020,Dimanov2020}.
Such results argue in favor of developing new explanation methods that are robust against such manipulation \citep{Alvarez-Melis2018,Hancox-Li2020}.
Initial research in this direction around robust optimization are promising \citep{Kulynych2020}, but their usefulness in practice remains to be seen.

\begin{challenge}
Too detailed an explanation can compromise a proprietary model that was intended to be kept confidential.
\end{challenge}
Research has shown that the output of feature importance methods like SHAP,
through repeated queries, are prone to membership attacks that
can reveal intimate details about the classification boundary \citep{Shokri2019}.
Similar research has shown that counterfactual explanations are
vulnerable to similar attacks \citep{Aivodji2020},
as are image-based explanations like saliency maps \citep{Zhao2021}.
Such results reveal the risk that when providing explanations to
external stakeholders,
the recipients of such explanations can collude to reconstruct
the inner workings of a model.
Such information leakage is not just theoretical,
but has been realized such as the public discovery of the Chase 5/24 rule,
simply by comparing decision outcomes across multiple applicants \citep{Kerr2021}.

\section{Conclusions and outlook}\label{sec:conclusions}

We have organized the XAI literature into seven distinct challenges
for providing a unified thesis for specifying the appropriate use of
XAI techniques to handle the myriad use cases in industry.
Numerous recent papers conclude that details such as the
specific intent and needs of multiple stakeholders,
global or local scope of desired explanation,
and type of data and model to be explained,
be they image classifiers, speech recognition engines,
adaptive chatbots, or recommender systems
matter greatly and argue in favor of customized techniques
for each application.
Furthermore, the lack of uniform evaluation criteria
for verifying the correctness of explanations necessitate a nuanced
consideration of how humans consumer and react to explanations provided to them,
and results showing tendencies to overtrust AI-generated explanations
ought to be factored in when crafting explainability guidelines for AI systems.
The innate brittleness of existing XAI techniques
means that they are vulnerable to malicious manipulation
to produce misleading evidence for reassuring overtrusting humans.
Finally, the desire for explanability needs to be balanced
against other competing needs such as privacy and security,
or else risk compromising the original AI system's intended purpose
or revealing details of a proprietary model that was intended to be kept confidential.
Our summary of the literature reveals these common themes
of real world challenges, thus meriting a cautious and principled approach
for using XAI in the full appreciation of the specific context of stakeholder needs,
business use case, and details of the AI system's construction.

\paragraph{Disclaimer}
This paper was prepared for informational purposes by the Artificial Intelligence Research  group of JPMorgan Chase \& Co and its affiliates (``JP Morgan''), and is not a product of the Research Department of JP Morgan. JP Morgan makes no representation and warranty whatsoever and disclaims all liability, for the completeness, accuracy or reliability of the information contained herein.  This document is not intended as investment research or investment advice, or a recommendation, offer or solicitation for the purchase or sale of any security, financial instrument, financial product or service, or to be used in any way for evaluating the merits of participating in any transaction, and shall not constitute a solicitation under any jurisdiction or to any person, if such solicitation under such jurisdiction or to such person would be unlawful. © 2021 JPMorgan Chase \& Co. All rights reserved.

\bibliographystyle{ACM-Reference-Format}
\bibliography{references}

\end{document}